\documentclass{article}
\usepackage{ijcai22}

\usepackage{times}
\usepackage{soul}
\usepackage{url}
\usepackage[hidelinks]{hyperref}
\usepackage[utf8]{inputenc}
\usepackage[small]{caption}
\usepackage{graphicx}
\usepackage{amsmath}
\usepackage{amsthm}
\usepackage{booktabs}
\usepackage{algorithm}
\usepackage{algorithmic}
\usepackage{amsfonts}
\usepackage{multirow}
\usepackage{color}

\title{Trustworthy Anomaly Detection: A Survey}

\author{
Shuhan Yuan $^1$ \and
Xintao Wu $^2$\\
\affiliations
$^1$Utah State University\\
$^2$University of Arkansas\\
\emails
shuhan.yuan@usu.edu,
xintaowu@uark.edu
}

\begin{document}

\maketitle

\begin{abstract}
    Anomaly detection has a wide range of real-world applications, such as bank fraud detection and cyber intrusion detection. In the past decade, a variety of anomaly detection models have been developed, which lead to big progress towards accurately detecting various anomalies. Despite the successes, anomaly detection models still face many limitations. The most significant one is whether we can trust the detection results from the models. In recent years, the research community has spent a great effort to design trustworthy machine learning models, such as developing trustworthy classification models. However, the attention to anomaly detection tasks is far from sufficient. Considering that many anomaly detection tasks are life-changing tasks involving human beings, labeling someone as anomalies or fraudsters should be extremely cautious. Hence, ensuring the anomaly detection models conducted in a trustworthy fashion is an essential requirement to deploy the models to conduct automatic decisions in the real world. In this brief survey, we summarize the existing efforts and discuss open problems towards trustworthy anomaly detection from the perspectives of interpretability, fairness, robustness, and privacy-preservation. 
\end{abstract}

\section{Introduction}
Anomaly detection aims to find instances that are deviated from normal ones. As a fundamental problem in machine learning with a wide spectrum of applications, anomaly detection has gained a lot of attention in recent years \cite{ruffUnifyingReviewDeep2021,pangDeepLearningAnomaly2020}. Although advanced anomaly detection techniques significantly improve the performance of identifying anomalies, there are still high risks to simply count upon the models in sensitive applications for automatic decision makings, such as health care or financial systems. Meanwhile, many anomaly detection tasks involve human beings, such as bank fraud detection, online troll detection, and malicious insider detection. Therefore, it is natural to ask whether we should trust algorithms for the task of labeling a person as an anomaly because every human being can be a potential victim of the algorithms due to intentional or unintentional misuse. 

Trustworthy AI, which emphasizes human value as principles in AI algorithms, such as fairness, transparency, accountability, reliability, privacy, and security, has become an urgent task for AI researchers \cite{liuTrustworthyAIComputational2021}. The main objective is to design trustworthy AI algorithms that address the social expectations of enhancing both AI ability and benefits to society without threat or risk of harm \cite{chengSociallyResponsibleAI2021,liuTrustworthyAIComputational2021}. 

In the context of anomaly detection, besides the effectiveness in detecting anomalies, trustworthy anomaly detection algorithms should also follow the typical social norms just like all the expectations to AI algorithms.
In this brief survey, we focus on four important dimensions of trustworthy anomaly detection, including interpretability, fairness, robustness, and privacy-preservation. 

\section{Brief Review to Anomaly Detection}
An anomaly is usually considered as an instance that deviates considerably from normality \cite{ruffUnifyingReviewDeep2021}. The anomaly detection model trained on a given dataset usually aims at deriving an anomaly score for an input test sample to distinguish anomalies from normal ones.

Due to the scarcity of anomalies in the training dataset, anomaly detection is usually conducted in unsupervised and  semi-supervised settings. The unsupervised anomaly detection setting usually assumes unlabeled data in the training set with rare anomalies, which follows the intrinsic property of data distributions in real-world situations. 
As anomalies are rare in real cases, one special case of the unsupervised setting, called one-class setting, further assumes all those unlabeled training data are normal samples.
On the other hand, the semi-supervised anomaly detection setting typically assumes a small number of labeled normal and abnormal samples and a large number of unlabeled samples in the training dataset. Recent studies have demonstrated that using a few anomalies for training can significantly improve anomaly detection performance \cite{ruffDeepSemiSupervisedAnomaly2020}. Because collecting anomalies is usually challenging, a special semi-supervised setting is positive-unlabeled learning, i.e., achieving anomaly detection based on labeled normal and unlabeled samples. 

Based on whether the complex deep neural networks (DNNs) are deployed, the anomaly detection models can be categorized into shallow and deep models. 
As anomaly detection has been studied for decades, many traditional machine learning models, i.e., shallow models, have been applied for detecting anomalies, such as support vector data description (SVDD) and principal component analysis (PCA). The shallow models are usually effective for detecting anomalies on tabular data, but hard for handling complex data types, such as text, image, or graph, due to lack of capability to capture non-linear relationships among features. In recent years, many deep learning models are proposed for anomaly detection because of their capability to automatically learn representations or patterns from a variety of data.
Both deep and shallow anomaly detection approaches can be mainly categorized into four groups, density estimation and probabilistic models, one-class classification models, reconstruction models, and other miscellaneous techniques \cite{ruffUnifyingReviewDeep2021}. 

\textbf{Density estimation and probabilistic models} detect anomalies by estimating the normal data distribution. The anomalies are detected with low likelihood in the normal data distribution. 
Generative models, including shallow models like Gaussian mixture models (GMMs) and deep models like variational autoencoder (VAE) and generative adversarial networks (GAN), have been applied for anomaly detection. Generative models  approximate the data distribution given a dataset and deep generative models are further able to model complex and high-dimensional data.

\textbf{One-class classification models} are discriminative approaches for anomaly detection, which  aim to learn a decision boundary based on normal samples. One-class support vector machine (OC-SVM)  as a classical one-class classification model is widely used for anomaly detection. Similarly, deep one-class classifiers such as Deep SVDD are developed recently to identify anomalies from complex data. 
Deep SVDD trains a neural network to map normal samples  enclosed to a center of the hypersphere in the embedding space and derives a sample's anomaly score based on its distance to the center.

\textbf{Reconstruction models} are trained to minimize reconstruction errors on the normal samples so that anomalies can be detected with large reconstruction errors. 
Reconstruction-based anomaly detection models include  traditional PCA,  which aims to find an orthogonal projection in a low-dimensional space with the minimum reconstruction error when mapping back to the original space, and recent deep autoencoder, which is parameterized by a deep neural network for encoding and decoding data.

\textbf{Other miscellaneous techniques}, which do not fall in the above three categories, are also developed for anomaly detection.  Isolation forest \cite{liuIsolationForest2008} builds trees from unlabeled samples and label anomalies those samples close to the root of trees. Local outlier factor (LOF), as another unsupervised anomaly detection approach,  computes the local density deviation of a given data point with respect to its neighbors, and labels points having a lower density than their neighbors as anomalies \cite{breunigLOFIdentifyingDensitybased2000}. As normal samples are usually easy to obtain, \cite{sippleInterpretableMultidimensionalMultimodal2020} adopts negative sampling approaches to generate anomalies and then leverages the classification models to achieve anomaly detection. Meanwhile, anomaly detection can be modeled as a special scenario of few-shot learning when a few anomalies are available for training \cite{pangExplainableDeepFewshot2021,shuhanyuanFewshotInsiderThreat2020}.

\begin{table*}[ht!]
\centering
\caption{Summary of Published Research in Trustworthy Anomaly Detection. (I--Interpretable; F--Fair; R--Robust; P--Privacy-preserving)}
\label{tb:summary}
\resizebox{.95\textwidth}{!}{%
\begin{tabular}{|l|l|l|l|l|l|l|l|l|}
\hline
                                      & Approach                                                                        & I         & F         & R         & P         & Structure & Data               & Key features                                              \\ \hline
\multirow{4}{*}{Density models}        & GEE \cite{nguyenGEEGradientbasedExplainable2019}               & $\bullet$ &           &           &           & VAE       & Multivariate       & Post-hoc gradient-based interpretation                    \\ \cline{2-9} 
                                    & OmniAnomaly \cite{suRobustAnomalyDetection2019}               & $\bullet$ &           &           &           & GRU+VAE       & Multivariate Time Series       & Post-hoc interpretation                    \\ \cline{2-9}
                                      & CAVGA  \cite{venkataramananAttentionGuidedAnomaly2020}         & $\bullet$ &           &           &           & VAE       & Image              & Intrinsic interpretation                                  \\ \cline{2-9} 
                                      & PPAD-CPS  \cite{keshkIntegratedFrameworkPrivacyPreserving2019} &           &           &           & $\bullet$ & GMM       & Multivariate       & Anonymization-based approach                              \\ \hline
\multirow{7}{*}{One-class models}      & FCDD \cite{liznerskiExplainableDeepOneClass2021}               & $\bullet$ &           &           &           & CNN       & Image              & Intrinsic interpretation                                  \\ \cline{2-9} 
                                      & EM-attention \cite{brownRecurrentNeuralNetwork2018}            & $\bullet$ &           &           &           & RNN       & Sequence           & Intrinsic interpretation                                  \\ \cline{2-9} 
                                      & AE-1SVM  \cite{nguyenScalableInterpretableOneclass2018}        & $\bullet$ &           &           &           & AE-OCSVM  & Multivariate/Image & Post-hoc gradient-based interpretation                    \\ \cline{2-9} 
                                      & OC-DTD  \cite{kauffmannExplainingAnomaliesDeep2020}            & $\bullet$ &           &           &           & OCSVM     & Image              & Post-hoc gradient-based interpretation                    \\ \cline{2-9} 
                                      & CutPaste  \cite{liCutPasteSelfSupervisedLearning2021}                                & $\bullet$ &           &           &           & CNN       & Image              & Post-hoc gradient-based interpretation                    \\ \cline{2-9} 
                                      & Deep Fair SVDD \cite{zhangFairDeepAnomaly2021}                 &           & $\bullet$ &           &           & CNN/MLP   & Multivariate/Image & Adversarial representation learning                       \\ \cline{2-9} 
                                      & Attack online AD \cite{kloftOnlineAnomalyDetection2010}        &           &           & $\bullet$ &           & SVDD      & Multivariate       & Poisoning attack                                          \\ \hline
\multirow{5}{*}{Reconstruction models} & AE Shapley  \cite{takeishiAnomalyInterpretationShapley2020}    & $\bullet$ &           &           &           & AE        & Multivariate       & Post-hoc perturbation-based interpretation                \\ \cline{2-9} 
                                      & DCFOD  \cite{songDeepClusteringBased2021}                      &           & $\bullet$ &           &           & MLP       & Multivariate       & Adversarial representation learning                       \\ \cline{2-9} 
                                      & FairOD \cite{shekharFairODFairnessawareOutlier2021}            &           & $\bullet$ &           &           & MLP       & Multivariate       & Fair regularizer                                          \\ \cline{2-9} 
                                      & APAE  \cite{goodgeRobustnessAutoencodersAnomaly2020}           &           &           & $\bullet$ &           & AE        & Multivariate       & Adversarial defense                                       \\ \cline{2-9} 
                                      & AE+PLS  \cite{loAdversariallyRobustOneclass2021}               &           &           & $\bullet$ &           & AE        & Multivariate       & Adversarial defense                                       \\ \hline
\multirow{6}{*}{Other Miscellaneous}  & NGIG  \cite{sippleInterpretableMultidimensionalMultimodal2020} & $\bullet$ &           &           &           & MLP       & Multivariate       & Post-hoc gradient-based interpretation \\ \cline{2-9} 
                                      & DevNet  \cite{pangExplainableDeepFewshot2021}                  & $\bullet$ &           &           &           & CNN       & Image              & Few-shot; post-hoc gradient-based interpretation          \\ \cline{2-9} 
                                      & FairLOF \cite{pFairOutlierDetection2020}                       &           & $\bullet$ &           &           & LOF       & Multivariate       & Heuristic principles                                      \\ \cline{2-9} 
                                      & PPLOF \cite{pFairOutlierDetection2020}                         &           &           &           & $\bullet$ & LOF       & Multivariate       & Cryptographic-based approach                              \\ \cline{2-9} 
                                      & Attack RNN-based AD \cite{xuApproachPoisoningAttacks2020}      &           &           & $\bullet$ &           & RNN       & Sequence           & Poisoning attack                                          \\ \cline{2-9} 
                                      & Robust AD via DP  \cite{duRobustAnomalyDetection2020}          &           &           & $\bullet$ & $\bullet$ & AE/RNN    & Image/Sequence     & Improve robustness via differential privacy               \\ \cline{2-9} 
                                      & DPSGD  \cite{abadiDeepLearningDifferential2016}             &            &               &        &  $\bullet$        &  DNN                  &    General          & General approach to achieve differential privacy          \\   \hline
\end{tabular}
}
\end{table*}

\section{Trustworthy Anomaly Detection}

From the trustworthy perspective, anomaly detection models should meet the following key properties.

\textbf{Performant:} Effectively detecting anomalies is the basic requirement for all anomaly detection models. Targeting on precisely identifying anomalies, anomaly detection models aim to reduce the rate of misclassifying normal samples as anomalies, or the rate of missing anomalies, or both. Meanwhile, due to the scarcity of anomalies, the classical classification evaluation metrics, such as accuracy, are not suitable to quantify the performance of anomaly detection. Commonly-used evaluation metrics for anomaly detection include area under the receiver operating characteristics (AUROC), area under the precision-recall curve (PRROC), and the false-alert rate when achieving 95\% of the true positive rate.  

\textbf{Interpretable:} When deploying black-box anomaly detection models to make critical predictions, there is an increasing demand for interpretability from various stakeholders \cite{preeceStakeholdersExplainableAI2018}.  For example, if a bank account is automatically suspended by algorithms due to potentially fraudulent activities, both its user and the bank would like to know explanation, e.g., what activities lead to the suspension. Therefore, interpretable anomaly detection requires the  models be able to provide interpretations to the detection results while maintaining the detection performance.

\textbf{Fair:} 
Ensuring the decisions free from discrimination against certain groups is critical when deploying anomaly detection models for high-stake tasks, e.g., predicting recidivism risk \cite{dressel2018accuracy}.
Any algorithmic bias against some groups in anomaly detection models should be avoided, which is a basic ethical requirement from the public.

\textbf{Robust:} Large pieces of work have demonstrated that machine learning models including anomaly detection models are vulnerable to various adversarial attacks. Model robustness is key to ensuring the reliability of models when deployed to real-world scenarios, like detecting outliers in roadways for intelligent transportation systems. Robustness in anomaly detection  requires the detection model has consistent outputs in face of attacking samples.

\textbf{Privacy-preserving:} Anomaly detection models require a large number of samples for training. As a result, massive data collection for model training raises privacy concerns. Properly protecting users' sensitive data is important and also required by government regulations. Privacy-preservation in anomaly detection should be able to protect the privacy of data in three aspects: data for training a model (input), the model itself, and the model's predictions (output). User private information should not be leaked in the whole pipeline of training and deploying an anomaly detection model. 

\begin{figure}[h]
    \centering
    \includegraphics[width=0.48\textwidth]{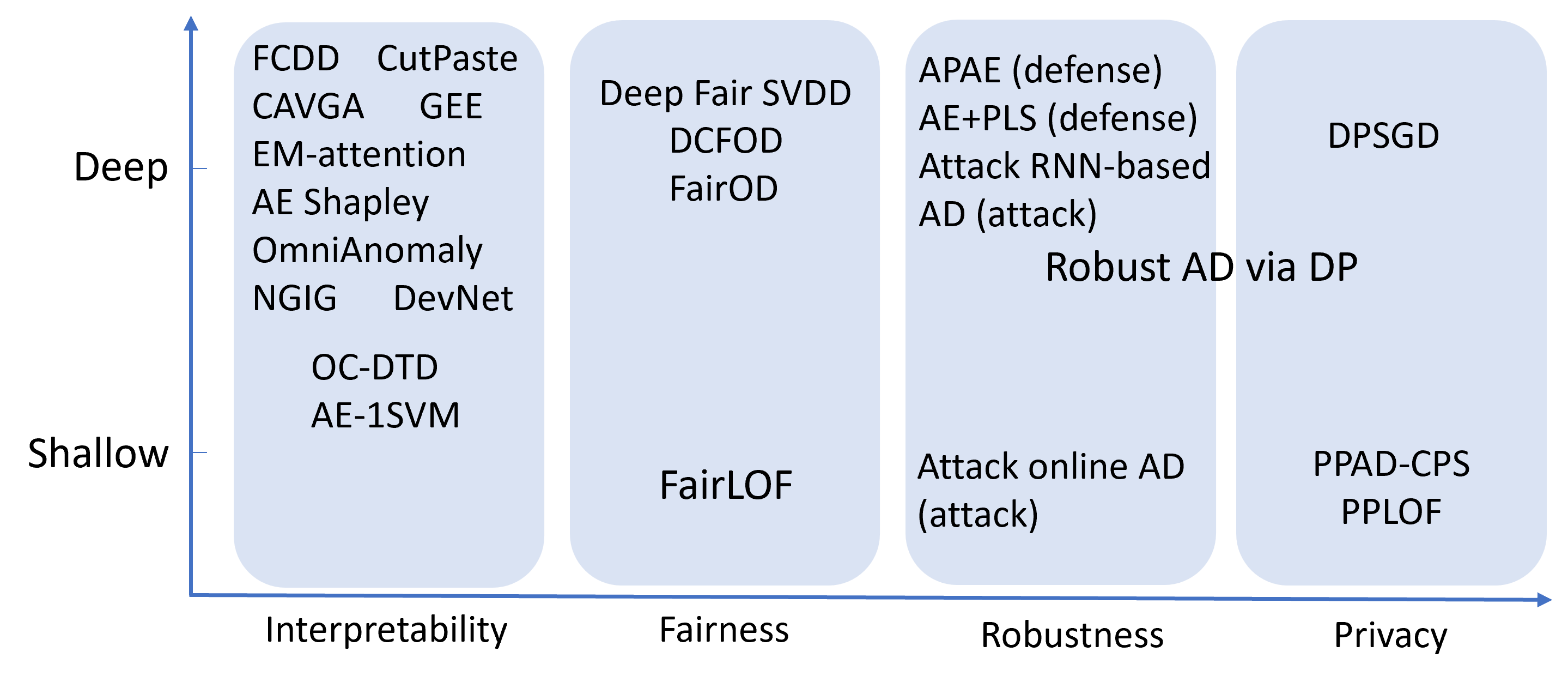}
    \caption{Summary of published research in trustworthy anomaly detection by using deep or shallow models.}
    \label{fig:summary}
\end{figure}

Table \ref{tb:summary} summarizes existing approaches towards trustworthy anomaly detection based on the above desideratum. Among them, the approach, Robust AD vis DP, addresses both robustness and privacy-preservation. Note that we skip the performance in this brief survey as several surveys in literature well cover this perspective \cite{ruffUnifyingReviewDeep2021,pangDeepLearningAnomaly2020}. Figure \ref{fig:summary} further summarizes the existing approaches towards trustworthy anomaly detection from the perspective of using deep or shallow models. Note that OC-DTD and AE-1SVM combine the shallow and deep models for interpretation.

\section{Interpretable Anomaly Detection}
Many state-of-the-art anomaly detection models are based on deep learning models that are considered as black-box models. Because anomaly detection models are commonly deployed in applications involving human beings, it is important to achieve interpretability to avoid potential risk against humans \cite{panjei2022survey}. 

Interpretable anomaly detection aims to provide explanations to individual or group predictions, especially for samples detected as anomalies. In literature, interpretable machine learning techniques can be grouped into two categories: intrinsic interpretability and post-hoc interpretability, depending on the time when the interpretability is achieved \cite{duTechniquesInterpretableMachine2018}. Intrinsic interpretability means the models are self-explanatory and usually is achieved by incorporating interpretability constraints or  adding interpretable layers in model structures. Post-hoc interpretability indicates the use of a second model to provide explanations for the deployed model. There are two commonly-used post-hoc interpretable techniques, perturbation-based and gradient-based approaches. Perturbation-based approaches try to find features that significantly affect model outcomes by sequentially removing or replacing a feature, while gradient-based approaches calculate the gradient of an output with respect to each input feature to derive its contribution

The majority of  existing interpretable machine learning models are developed for supervised learning tasks, such as image and text classification tasks, where the labels are available. However, anomaly detection models are usually semi-supervised or unsupervised learning models. How to interpret  anomaly detection results with only observing the normal samples is a challenge. 

\subsection{Approaches}
Several studies extend the density-based anomaly detection approaches to achieve interpretability. GEE adopts VAE to capture normal patterns and derives anomaly scores for multivariate input samples \cite{nguyenGEEGradientbasedExplainable2019}. The post-hoc interpretation is achieved based on a gradient-based approach, where features with large absolute gradients indicate significant contribution to the output. CAVGA adopts the convolution-VAE to learn the latent representation with the preservation of spatial information of images and incorporates an attention map for interpretation \cite{venkataramananAttentionGuidedAnomaly2020}. The attention map is trained to cover all the normal regions and the anomalous regions can then be highlighted as the attention map without covering.

Several interpretable one-class anomaly detection models are developed.
FCDD provides intrinsic interpretation by generating anomaly heat maps to highlight anomalous regions in abnormal maps \cite{liznerskiExplainableDeepOneClass2021}. Its core idea is to generate a matrix with entries corresponding to the abnormal regions in an image. If an image is abnormal, the sum of all entries in the matrix should be maximized, while the sum for a normal image  should be minimized. Besides image data, an attention mechanism, which can highlight the important features in sequential data, is used to identify anomalous events in log anomaly detection \cite{brownRecurrentNeuralNetwork2018}. 
To enhance the traditional one-class anomaly detection models with explanations, one direction is neuralization, which is to convert the non-neural network models to functionally equivalent neural network models so that the interpretation techniques developed for deep learning models can be leveraged \cite{kauffmannExplainingAnomaliesDeep2020}. For example, OC-DTD models one-class SVM as a three-layer structure, consisting of feature extraction, distance computation, and pooling, and then applies the Layer-wise Relevance Propagation (LRP) interpretation approach to assign scores to  input features \cite{kauffmannExplainingAnomaliesDeep2020}. Another direction is to combine the deep model with the  shallow model. 
\cite{nguyenScalableInterpretableOneclass2018} develops an autoencoder based one-class support vector machine (AE-1SVM) and uses the gradient-based attribution methods to explain the detection results.

Regarding the reconstruction-based anomaly detection models,  \cite{antwargExplainingAnomaliesDetected2020} develops interpretable autoencoder models to identify features leading to high reconstruction errors based on SHAP, which is able to provide post-hoc interpretation of detection results.  

Some other techniques are also developed to achieve interpretable anomaly detection. To leverage the interpretable techniques for supervised learning models, \cite{sippleInterpretableMultidimensionalMultimodal2020} develops a framework, which first generates anomalies by negative sampling to train a binary classifier and then adopts a gradient-based approach to detect abnormal features. InterpretableSAD further extends the above idea to achieve interpretable sequence anomaly detection \cite{han2021interpretablesad}. DevNet is a few-shot learning-based anomaly detection model and also leverages the idea of the gradient-based approach to detect abnormal regions in an image \cite{pangExplainableDeepFewshot2021}.

\subsection{Discussions}
Existing approaches mainly focus on providing interpretation on point anomalies, such as abnormal pixels in an image. Providing explanations on contextual and collective anomalies is more challenging and greatly needed. Meanwhile, besides providing local interpretation for a single prediction result,  understanding why a group of samples are labeled as anomalies, i.e., local interpretability for a group of predictions, can provide more insights into abnormal behaviors.  

\section{Fair Anomaly Detection}
The research community has well recognized that data-driven decision algorithms can produce unfair or biased decisions against a group of people, especially the minority groups, on the basis of some personal characteristics, such as gender, race, or age, called protected or sensitive attributes. Therefore, algorithmic fairness has received substantial attention in recent years \cite{zhangAntidiscriminationLearningCausal2017,mehrabiSurveyBiasFairness2021}. Currently, a large number of studies aim to achieve fairness in classical machine learning tasks, such as classification, regression, clustering, or dimensionality reduction \cite{mehrabiSurveyBiasFairness2021}. Recently, researchers notice that ensuring fairness in anomaly detection is also critical. However, achieving fair anomaly detection has its unique challenges due to the potentially high correlation between minority groups and outliers, which means that anomaly detection approaches could produce injustice outcomes by overly flagging the samples from the minority groups as outliers. Currently, only a few studies target this emerging task. 

\subsection{Approaches}
Following the typical requirements for fair machine learning models, existing approaches usually decompose fair anomaly detection into two objectives, anomaly detection and fair learning. Therefore, fair anomaly detection aims to find potential anomalies that substantially differ from the majority instances while maintaining insignificant to sensitive attribute subgroups  \cite{davidsonFrameworkDeterminingFairness2020}.

A few notions of fair anomaly detection are proposed, such as treatment parity or statistical parity \cite{shekharFairODFairnessawareOutlier2021,songDeepClusteringBased2021}. The basic idea is to ensure the anomaly detection result is invariant or independent of the protected attributes. Therefore, fairness  can be evaluated based on the detection difference on protected and unprotected groups in terms of an anomaly detection metric such as AUC.  

FairLOF \cite{pFairOutlierDetection2020} is one pioneering work to tackle the fair anomaly detection problem with the focus on the local outlier factor (LOF) algorithm. FairLOF proposes several heuristic principles to mitigate unfairness via incorporating a correction term on LOF's distance function  so that samples in under-represented groups have a similar distance distribution with the whole population.

In the category of one-class anomaly detection, Deep Fair SVDD enhances the Deep SVDD model with a fairness guarantee \cite{zhangFairDeepAnomaly2021}. 
The approach adopts an adversarial fair representation learning model such that a discriminator cannot predict the protected attribute given representations derived by Deep SVDD. Therefore, the anomaly detection results are independent of the protected attribute.

In the category of reconstruction-based anomaly detection,  two approaches based on autoencoder are developed. DCFOD adopts adversarial training to ensure the representations derived from autoencoder free from the protected attributes' information with the discriminator's guidance \cite{songDeepClusteringBased2021}.  FairOD incorporates fairness regularization terms in addition to the regular reconstruction loss into the objective function to minimize the absolute correlation between the reconstruction error and protected attributes \cite{shekharFairODFairnessawareOutlier2021}.

\subsection{Discussions}
Currently,  research on fair anomaly detection is still in its infancy. In the literature of fair machine learning, researchers have proposed a large number of fairness notions in categories of group fairness, sub-group fairness, and individual fairness. Current fair anomaly detection approaches only achieve group fairness in terms of demographic parity that requires equal detection rates across groups.  Other fairness notions, especially individual fairness, are still under-exploited. Meanwhile, as minority groups could have a strong correlation with anomalies, correlation-based anomaly detection approaches could mislabel minority groups as anomalies. It is imperative to develop causation-based anomaly detection approaches that can provide a way to achieve causation-based fair anomaly detection.
    
\section{Robust Anomaly Detection}
Recent studies have demonstrated that machine learning models are vulnerable to intentionally-designed input samples from attackers, which can lead to catastrophic results.

There are two common attacking strategies: 1) data poisoning attacks that are conducted on the training time by injecting poisoned samples into training data to compromise the learning model; 2) evasion attacks that try to fool the detection models to make incorrect predictions on the testing phase by carefully designed samples \cite{chakrabortyAdversarialAttacksDefences2018,zhangAdversarialAttacksDeeplearning2020}. Meanwhile, attacks can be further categorized into white-box attacks or black-box attacks based on the amount of information available to attackers. 
In white-box attacks, the attacker has complete access to the model, including its structure and weights, and in black-box attacks, the attacker can only access the probabilities or the labels of some probed inputs.

One connection between adversarial attacks and anomaly detection is that adversarial samples can be considered as extremely  hard-to-detect out-of-distribution samples \cite{bulusuAnomalousExampleDetection2021,ruffUnifyingReviewDeep2021} because these samples are crafted to mislead learning models. Several studies have demonstrated that out-of-distribution detection can improve the model robustness against adversarial attacks
\cite{hendrycksUsingSelfSupervisedLearning2019}. 
While it is interesting to study model robustness through out-of-distribution detection, in this paper, we focus on reviewing adversarial attacks and defenses for anomaly detection itself, where the attacker aims to craft adversarial samples to evade detection or corrupt detection models. 

\subsection{Approaches}
Some traditional one-class anomaly detection models, such as PCA, are sensitive to  noisy samples in the training set. Some outliers in the normal training set may totally destroy the performance of detection models. Therefore, robust models, such as robust PCA or robust SVM, are developed to make detection models robust to noisy data that are arbitrarily corrupted.
\cite{kloftOnlineAnomalyDetection2010} studies the poisoning attacks in an online one-class anomaly detection scenario, where an anomaly score is derived based on a data point's distance to the center of normal samples. The goal of the attack is to make the detection model accept anomalies with large distances to the center. The theoretical analysis shows that if the attacker can control the training data more than a traffic ratio, the poisoning attack can succeed to push the center towards the attacking points. On the other hand, if the attacker can control only a small ratio of training data, the attack would fail with an infinite effort.

Recent studies have demonstrated that many popular white-box adversarial attack approaches can be adapted for attacking anomaly detection models, especially the reconstruction-based anomaly detection models, while common defense approaches for classifiers are not very effective for anomaly detection models \cite{loAdversariallyRobustOneclass2021,goodgeRobustnessAutoencodersAnomaly2020}.  
For example, the autoencoder-based anomaly detection models are vulnerable to many white-box attacking approaches, such as fast gradient sign method (FGSM) \cite{kurakinAdversarialMachineLearning2017} or projected gradient descent (PGD) \cite{madryDeepLearningModels2018}. As a result, perturbations on anomalous samples can cause the model to misclassify the anomalies as normal samples by making the reconstruction errors decrease rather than increase. To defend against such attacks, \cite{goodgeRobustnessAutoencodersAnomaly2020} develops an approximate projection autoencoder (APAE) model with the combination of two defense strategies, approximate projection and feature weighting. The approximate projection is to update the latent embeddings derived from the encoder to ensure the reconstructions have minimum changes, while the feature weighting is to normalize reconstruction errors to prevent the anomaly score from being dominated by the poorly reconstructed features. 
\cite{loAdversariallyRobustOneclass2021} also applies the idea of updating latent embeddings and develops Principal Latent Space (PLS), which purifies the latent embeddings of autoencoder based on principal component analysis.

\subsection{Discussions}
From the attack perspective, there are very limited studies in literature focusing on the adversarial attack against anomaly detection models. Recently, \cite{xuApproachPoisoningAttacks2020} develops a poisoning attack approach against RNN-based anomaly detection models and shows the attack can  generate discrete adversarial samples to damage the RNN model with a few instances. Various adversarial attack strategies, such as exploratory attacks that try to gain information about the model states by probing the models \cite{chakrabortyAdversarialAttacksDefences2018} or backdoor attacks that create a backdoor into a learning model \cite{chenTargetedBackdoorAttacks2017}, are not explored in the anomaly detection context. It is imperative to know how effective these potential adversarial attacks would be against anomaly detection models before they are launched by attackers in real-world applications.

From the defense perspective, existing studies focus on model-specific defending strategies for whit-box attacks. Other defending strategies are also worth exploring. For example, adversarial training is a widely-used strategy to improve model robustness against adversarial attacks by incorporating adversarial samples in the training set \cite{goodfellowExplainingHarnessingAdversarial2015}. 
However, under the settings of unsupervised or one-class learning, we do not have anomalies. In such cases, how to craft adversarial samples, especially the anomalies in the purpose of evading detection, is challenging. 
It is also interesting to develop new anomaly detection approaches to detect adversarial samples that are crafted for anomaly detection models, e.g., based on the idea of using out-of-distribution detection models to detect adversarial samples from a polluted training set \cite{bulusuAnomalousExampleDetection2021}.

\section{Privacy-preserving Anomaly Detection}
Many anomaly detection tasks involve the use of personal data, such as online troll detection, bank fraud detection, or insider threat detection. 
Therefore, privacy-preservation as a process of protecting sensitive information against being revealed or misused by unauthorized users has been studied extensively \cite{xu2021privacy}. The goal of privacy-preserving anomaly detection is to achieve anomaly detection with the protection of sensitive information and without compromising its effectiveness.

\subsection{Approaches}
Privacy-preserving anomaly detection can be categorized into three classes, anonymization-based, cryptographic-based, and perturbation-based approaches. 

Anonymization-based approaches are to achieve privacy protection by removing personally identifiable information while keeping the utility of the data for anomaly detection. 
\cite{keshkIntegratedFrameworkPrivacyPreserving2019} develops a privacy-preserving anomaly detection framework for cyber-physical systems (PPAD-CPS). Its key idea is to filter and transform original data into a new format in the data-processing step, and adapt the Gaussian Mixture Model and Kalman Filter to detect anomalies on the anonymized data.

Privacy-preserving LOF (PPLOF) is a cryptographic-based approach for anomaly detection over vertically partitioned datasets among two parties \cite{liPrivacypreservingLOFOutlier2015}. 
A privacy-preserving windowed Gaussian anomaly detection algorithm on encrypted data is developed for an edge computing environment in  \cite{mehnazPrivacypreservingRealtimeAnomaly2020}.

Differential privacy (DP) is the de facto standard for conducting perturbation-based privacy-preserving data analysis \cite{dworkAlgorithmicFoundationsDifferential2014}.
However, fraud detection and differential privacy protection are intrinsically conflicting  because the goal of DP is to conceal presence or absence of any particular instance and the problem of fraud detection is to identify anomaly instances. \cite{DBLP:conf/pkdd/OkadaFS15} balances these two tasks by using two differentially private queries, counting the number of outliers in a subspace and discovering top subspaces with a large number of outliers,  to understand behavior of outliers. DP can generally be achieved by injecting some random noises to model's input data, output, or objective function. To ensure differential privacy in fraud detection, we can collect input data with local differential privacy protection, use  differentially private aggregates or statistics for downstream anomaly detection tasks \cite{fanDifferentiallyPrivateAnomaly2013}, or train deep learning based anomaly detection models by differentially private stochastic gradient descent (DPSGD) \cite{abadiDeepLearningDifferential2016}, which adds Gaussian noise to the aggregated gradients during the gradient descent process.

\subsection{Discussions}
There is always a trade-off between privacy protection and other quality of models, such as model effectiveness or efficiency. It is well-known that cryptographic-based approaches require high computational overhead, and perturbation-based approaches could damage the model's accuracy. 
Recently \cite{duRobustAnomalyDetection2020} shows that applying differential privacy can potentially improve the robustness of anomaly detection when the training set contains noises. The intuition is that deep learning models with millions of parameters are capable of remembering all the training samples including the noises. In the unsupervised anomaly detection scenario, the model would consider rare samples, such as backdoor samples, in the training set as normal ones due to potential overfitting to rare samples. However, DP makes the model underfit rare samples by injecting random noise during the training so that the overall performance can be improved.
It is worth studying in what circumstances privacy-preserving approaches can also improve performance and robustness.

\section{Future Directions}
In this brief survey, we review the current research in trustworthy anomaly detection from four perspectives, interpretability, fairness, robustness, and privacy-preservation. Overall, compared with widely-studied classification models, ensuring trustworthiness of anomaly detection models still has a large room for improvement. Below we point out several potential future directions.

{$\bullet$ \bf \noindent Benchmark Datasets.}
There are limited datasets designed for trustworthy anomaly detection. For interpretable anomaly detection, only a few image and time series datasets with fine-grained labels regarding abnormal regions can be used for evaluation \cite{bergmannMVTecADComprehensive2019,jacobExathlonBenchmarkExplainable2021}. Meanwhile, for fair anomaly detection, current studies usually adopt widely-used datasets in the fair machine learning domain, such as COMPAS \cite{dressel2018accuracy}, but these datasets are not designed for anomaly detection. Therefore, developing datasets related to trustworthy anomaly detection is critical for conducting rigorous evaluation and boosting this line of research.

{$\bullet$  \bf \noindent Model Agnostic Approaches.}
Current approaches to achieve specific trustworthy desideratum for anomaly detection are often developed for each specific model, such as fair autoencoder or interpretable CNN. Different anomaly detection models are developed to tackle different types of data such as images, videos, audios, text, graphs, and sequences,  developing model-agnostic approaches for trustworthy anomaly detection is important yet exploited.

{$\bullet$  \bf \noindent Achieving Multiple Desiderata.} 
Interpretability, fairness, robustness, and privacy are not orthogonal to each other. 
Several recent works have studied their connections in learning. Note that one desideratum would affect another in an either positive or negative manner. For example, \cite{bagdasaryanDifferentialPrivacyHas2019} shows that differential privacy can lead to disparate impact against minority groups and \cite{DBLP:conf/kdd/XuDW21} studies how to mitigate the disparate impact. On the other hand, \cite{duRobustAnomalyDetection2020} shows differential privacy improves model robustness against backdoor attacks. Moreover, privacy-preserving techniques can also act as defenders to protect models from some other attacks, such as membership inference attacks. \cite{xuBeRobustBe2021} shows adversarial learning based defense techniques do not provide sufficient protection to minority groups, which incurs unfairness. On the contrary, \cite{liuAdversarialAttacksDefenses2021} shows adversarial attack and defense can help interpretation because both adversarial attack and interpretation need to identify important features of input samples or important parameters of models. 
In addition to understanding their inherent relationships, it is imperative to develop a unified anomaly detection framework and new mechanisms that can achieve multiple trustworthy desiderata simultaneously. For example, \cite{DBLP:conf/www/XuYW19} shows both differential privacy and fairness can be achieved by adding less noise to the objective function of logistic regression than adding noise separately. Moreover, in addition to four desiderata studied in this survey, it is also important to study other dimensions such as  auditability, and environmental well-being \cite{liuTrustworthyAIComputational2021} in anomaly detection. However, there are very limited research in this direction.

\clearpage

\end{document}